\documentclass[journal]{IEEEtran}

\ifCLASSINFOpdf
\else
   \usepackage[dvips]{graphicx}
\fi
\usepackage{url}
\usepackage{amsmath,graphicx,amsfonts,algorithm,algorithmic,amsopn,amssymb,times,hyperref,subfigure}
\begin{document}
\title{Robust Dynamic Multi-Modal Data Fusion: \\A Model Uncertainty Perspective}
\author{Bin Liu
\thanks{First posted on May 12th, 2021. Revised on July 8th, 2021.}
\thanks{Bin Liu is with Zhejiang Lab, Hangzhou, 311100 China (e-mail: bins@ieee.org).}}
\markboth{}
{Shell \MakeLowercase{\textit{et al.}}: Bare Demo of IEEEtran.cls for IEEE Journals}
\maketitle
\begin{abstract}
This letter is concerned with multi-modal data fusion (MMDF) under unexpected modality failures in nonlinear non-Gaussian dynamic processes. An efficient framework to tackle this problem is proposed. In particular, a notion termed modality ``\emph{usefulness}", which takes a value of 1 or 0, is used for indicating whether the observation of this modality is useful or not. For $n$ modalities involved, $2^n$ combinations of their ``\emph{usefulness}" values exist. Each combination defines one hypothetical model of the true data generative process. Then the problem of concern is formalized as a task of nonlinear non-Gaussian state filtering under model uncertainty, which is addressed by a dynamic model averaging (DMA) based particle filter (PF) algorithm. This DMA algorithm employs $2^n$ models, while all models share the same state-transition function and a unique set of particle values. That makes its computational complexity only slightly larger than a single model based PF algorithm, especially for scenarios in which $n$ is small. Experimental results show that the proposed solution outperforms remarkably state-of-the-art methods. Code and data are available at \url{https://github.com/robinlau1981/fusion}.
\end{abstract}
\IEEEpeerreviewmaketitle
\section{Introduction}
\IEEEPARstart{M}{ulti-modal} data fusion (MMDF) is used for analyzing data collected from multiple data acquisition frameworks (DCFs). Each type of DCF is associated with a data modality. Here we consider a crucial concern related to MMDF, namely how to make the MMDF result immune to unexpected modality failures. Robust DF methods exist, while they only consider co-modal data or static linear Gaussian models, see e.g., \cite{kumar2007method,chen2012robust}. There lacks a general and effective solution to deal with cross-modal DF in nonlinear non-Gaussian dynamic processes. It is the setting that motivates this work. Compared with co-modal DF, cross-modal DF is much more difficult to handle. For co-modal DF, an explicit correlation relationship among the measurements is available for use in monitoring modality failures. In the context of cross-modal DF, what links data of different modalities is the unknown system state that needs to be estimated, and thus no explicit correlation can be employed. The nonlinear, non-Gaussian, and dynamic property of the system of our concern further exacerbates the difficulty of the problem.

In this paper, we propose a novel model uncertainty based point of view to address the aforementioned challenge. We show that by introducing a notion termed modality ``\emph{usefulness}", and then enumerating possible values of the ``\emph{usefulness}" vector, we can construct a set of candidate models characterizing all possible forms of the data generative process. In particular, based on the above perspective, we propose a DMA based MMDF algorithm and demonstrate its remarkable performance benefit compared with existent state-of-the-art methods.

From a computation perspective, our algorithm has straightforward connections to robust particle filtering (RPF) methods in \cite{liu2017robust,liu2011instantaneous,liu2018ilapf,liu2019robust,liu2020data,qi2019dynamic,el2021particle,song2020particle,urteaga2016sequential}. The difference lies in that such RPF methods focus on how to handle uncertainties underlying the state-transition function or how to model the observation noise. In contrast, our algorithm presented here addresses the uncertainty underlying the ideal way in which the cross-modal observations are fused. The term ``ideal way" means the way that is consistent with the true data generating mechanism, which is also the way that leads to the most accurate state estimations. Our method also has a relationship in spirit with trust models \cite{cho2015survey,liu2020survey}. In particular, if we take the ``\emph{usefulness}" value as a trust metric, then our method can be regarded as a probabilistic trust model. Existent trust models consider co-modal data generated in linear Gaussian systems (see e.g., \cite{wang2017online,liu2015toward,liu2017state}), or focus on fusion of information
generated from multiple sources including human in a static setting \cite{pong2007empirical,nevell2010fusion,josang2012interpretation,stampouli2010fusion}. In contrast, our model presented here can deal with cross-modal DF in the context of nonlinear non-Gaussian dynamic systems.

The main contribution of this paper is threefold. First, we provide a model uncertainty based perspective to view the problem MMDF. Using this perspective, we propose a model for cross-modal DF in nonlinear non-Gaussian dynamic
systems (Section \ref{sec:proposed_model}). Then we derive a corresponding algorithm based on that model (Section \ref{sec:algorithm}). Finally, we test our algorithm based on comprehensive experiments and make our code and data open source at \url{https://github.com/robinlau1981/fusion}. Results show a remarkable performance advantage of our algorithm over state-of-the-art methods (Section \ref{sec:experiment}).
\section{Model}
In this section, we present our model for robust MMDF.
To begin with, we introduce a general model to fix notations.
For the sake of clarity in presentation, we only consider 2 modalities here, while our model and algorithm can be easily extended to deal with more modalities. Consider a state space model (SSM) defined by a state-transition prior density  $p(\textbf{x}_t|\textbf{x}_{t-1})$ and a likelihood function
\begin{equation}\label{eqn:likelihood}
L_{t}(\textbf{x}_t)\triangleq p(\textbf{y}_{1,t},\textbf{y}_{2,t}|\textbf{x}_t)
\end{equation}
where $t$ denotes the discrete-time index, $\textbf{x}\in\mathbb{R}^{d_x}$ the hidden state to be estimated,
$\textbf{y}_{i,t}\in\mathbb{R}^{d_i}$ the observation of the $i$th modality at time $t$, $d_i$ the dimension of $\textbf{y}_{i,t}$, $i=1,2$.
We make an appropriate assumption that $\textbf{y}_{1,t}$ and $\textbf{y}_{2,t}$ are independent given $\textbf{x}_t$, then Eqn. (\ref{eqn:likelihood}) factorizes as follows
\begin{equation}\label{eqn:likelihood_factor}
L_{t}(\textbf{x}_t)\triangleq L_{1,t}(\textbf{x}_t)L_{2,t}(\textbf{x}_t),
\end{equation}
where $L_{i,t}(\textbf{x}_t)\triangleq p(\textbf{y}_{i,t}|\textbf{x}_t)$ denotes the likelihood function associated with the $i$th modality, $i=1,2$.

The task is to calculate the posterior probabilistic density function (pdf) of $\textbf{x}_t$, denoted by $p_{t|t}\triangleq p(\textbf{x}_t|\textbf{y}_{1,1:t}, \textbf{y}_{2,1:t})$, where $\textbf{y}_{i,1:t}\triangleq\left[\textbf{y}_{i,1},\ldots,\textbf{y}_{i,t}\right]$. Based on Bayesian theorem, $p_{t|t}$ can be computed from $p_{t-1|t-1}$ recursively as follows
\begin{equation}\label{eqn:filter}
p_{t|t}=\frac{L_{t}(\textbf{x}_t)\int p(\textbf{x}_{t}|\textbf{x}_{t-1})p_{t-1|t-1}d\textbf{x}_{t-1}}{p(\textbf{y}_{1,1:t}, \textbf{y}_{2,1:t}|\textbf{y}_{1,1:t-1}, \textbf{y}_{2,1:t-1})}.
\end{equation}
For nonlinear non-Gaussian cases, there is no analytical solution to Eqn.(\ref{eqn:filter}), while one can use PF to obtain an approximated solution \cite{arulampalam2002tutorial}.
\subsection{The Proposed Model for robust MMDF}\label{sec:proposed_model}
The model presented above assumes that each modality works normally as expected at each time step. Here we extend it to handle unexpected modality failures.
To begin with, we introduce a notion termed modality ``\emph{usefulness}", denoted by $\mathcal{U}\in\{1,0\}$. We denote the ``\emph{usefulness}" of the $i$th modality at time $t$ by $\mathcal{U}_{i,t}$, and use $\mathcal{U}_{i,t}=1 (0)$ to represent the hypothesis that the observation of the $i$th modality at time $t$ is useful (useless) for estimating $\textbf{x}_{t}$.

Recall that we consider 2 modalities here for the sake of clarity in presentation. The combination of ``\emph{usefulness}" values of 2 modalities can be represented by a vector $[\mathcal{U}_{1,t}, \mathcal{U}_{2,t}]$. This ``\emph{usefulness}" vector has 4 value options in total, namely $[1,1],[1,0],[0,1]$ and $[0,0]$. Each value option corresponds to a specific hypothesis of the true likelihood function $L^{\star}_{t}(\textbf{x}_t)$. Specifically, the value option $[1,1]$ corresponds to a hypothesis of $L^{\star}_{t}(\textbf{x}_t)=L^{(1)}_{t}(\textbf{x}_t)\triangleq L_{1,t}(\textbf{x}_t)L_{2,t}(\textbf{x}_t)$, the same as in Eqn.(\ref{eqn:likelihood_factor}); $[1,0]$ to $L^{\star}_{t}(\textbf{x}_t)=L^{(2)}_{t}(\textbf{x}_t)\triangleq L_{1,t}(\textbf{x}_t)L^0_2$; $[0,1]$ to $L^{\star}_{t}(\textbf{x}_t)=L^{(3)}_{t}(\textbf{x}_t)\triangleq L^0_1L_{2,t}(\textbf{x}_t)$;and $[0,0]$ to $L^{\star}_{t}(\textbf{x}_t)=L^{(4)}_{t}(\textbf{x}_t)\triangleq L^0_1L^0_2$. Here $L^0_i$ denotes the likelihood function of the $i$th modality in case of its observation being useless, due to e.g., sensor faults or communication failures.

Assume that, when the $i$th modality fails at time $t$, $\textbf{y}_{i,t}$ is uniformly distributed across its value space, whose volume is denoted by $V_i$. Set $L^0_i=1/V_i$ if $\textbf{y}_{i,t}$ is within this value space, otherwise let $L^0_i=0$. Note that this is an appropriate likelihood function for doing Bayesian model selection or averaging, as it integrates to 1 over the value space of $\textbf{y}_{i,t}$.

Now we have 1 state transition prior density function $p(\textbf{x}_t|\textbf{x}_{t-1})$ and 4 potential likelihood functions, namely $L^{(m)}_{t}(\textbf{x}_t), m=1,\ldots, 4$. Each pair of $p(\textbf{x}_t|\textbf{x}_{t-1})$ and $L^{(m)}_{t}(\textbf{x}_t)$ constitute a candidate model, denoted by $\mathcal{M}_m$, which defines a specific data generating mechanism at time $t$. Following \cite{liu2017robust}, let $\mathcal{H}_t=m$ denote the event that the $m$th candidate model captures the real data generative mechanism at time $t$. Then we can derive the posterior pdf under this multi-model setting based on the Bayesian model averaging theory \cite{hoeting1999bayesian,raftery1997bayesian}:
\begin{equation}\label{eqn:bma}
p_{t|t}=\sum_{m=1}^4p_{m,t|t}\pi_{m,t|t},
\end{equation}
where $p_{m,t|t}\triangleq p(\textbf{x}_t|\mathcal{H}_t=m,\textbf{y}_{1,1:t}, \textbf{y}_{2,1:t})$, $\pi_{m,t|t}\triangleq p(\mathcal{H}_{t}=m|\textbf{y}_{1,1:t},\textbf{y}_{2,1:t})$ and $\pi_{m,t|t}$ can be seen as the weight of $\mathcal{M}_m$ in $p_{t|t}$, $m=1,\ldots,4$.
\section{Algorithm}\label{sec:algorithm}
In this section, we derive a dynamic model averaging (or DMA in short) algorithm for robust MMDF based on the model proposed in subsection \ref{sec:proposed_model}.

We consider a recursive solution to approximate Eqn.(\ref{eqn:bma}) online under the PF framework. The algorithm is initialized by specifying a weighted particle set $\{x_{1}^i,\omega_{1}^i\}_{i=1}^N$ that satisfies $0<\omega_{1}^i<1, \forall i, \sum_{i=1}^N\omega_{1}^i=1$, $p_{1|1}\simeq\sum_{i=1}^N\omega_{1}^i\delta_{x_{1}^i},$ and setting $\pi_{m,1|1}=1/M$, $m=1,\ldots,M$. Here $N$ denotes the particle size, $\delta_{x}$ the Dirac-delta function located at $x$, and $M$ the number of candidate models ($M=4$ as shown in subsection \ref{sec:proposed_model}). We use $x$ and $y$ to denote realizations of $\textbf{x}$ and $\textbf{y}$, respectively.

At time $t, t>1$, the task is to approximate Eqn.(\ref{eqn:bma}) based on the available information encoded by $\{x_{1:t-1}^i,\omega_{t-1}^i\}_{i=1}^N$ and $\pi_{m,t-1|t-1}$, $m=1,\ldots,M$.
First, we perform one-step state transition for each particle. Specifically, we draw $\hat{x}_t^i$
from the state-transition prior $p(\textbf{x}_t|x^i_{t-1}), \forall i$, following the same spirit of the bootstrap filter \cite{gordon1993novel}. The particle weights associated with the $m$th candidate model are calculated as below,
\begin{equation}\label{eqn:omega_mm}
\omega_{m,t}^i=\frac{\omega_{t-1}^iL^{(m)}_{t}(\hat{x}_t^i)}{\sum_{j=1}^{N}\omega_{t-1}^jL^{(m)}_{t}(\hat{x}_t^j)}.
\end{equation}
According to the theory of importance sampling, $p_{m,t|t}$ can then be approximated as follows
\begin{equation}\label{eqn:posterior_m_approx}
p_{m,t|t}\simeq\sum_{i=1}^N\omega_{m,t}^i\delta_{\hat{x}_{t}^i}.
\end{equation}

Now let consider how to derive $\pi_{m,t|t}$ from $\pi_{m,t-1|t-1}$, given observations $y_{1,t}$ and $y_{2,t}$.
First, we specify an appropriate hypothesis transition process for the algorithm agent to predict the value of $\mathcal{H}$ at time $t$ before it seeing $y_{1,t}$ and $y_{2,t}$. In particular, we let
\begin{equation}
\pi_{m,t|t-1}=\pi_{m,t-1|t-1}.
\end{equation}
Then, based on the Bayesian rule, we have
\begin{equation}\label{posterior_model_indicator}
\pi_{m,t|t}=\frac{\pi_{m,t|t-1}g_{m,t|t-1}}{\sum_{m=1}^M\pi_{m,t|t-1}g_{m,t|t-1}},
\end{equation}
where $g_{m,t|t-1}\triangleq p_m(y_{1,t},y_{2,t}|y_{1,1:t-1},y_{2,1:t-1})$ denotes the marginal likelihood of
$\mathcal{M}_m$ at time $t$, which is shown to be
\begin{equation}\label{marginal_lik}
g_{m,t|t-1}=\int L^{(m)}_{t}(x_t)p(x_t|y_{1,1:t-1},y_{2,1:t-1})dx_t.
\end{equation}
The above integral can be approximated as follows
\begin{equation}\label{eqn:particle_marignal_lik}
g_{m,t|t-1}\simeq\sum_{i=1}^N \omega_{t-1}^iL^{(m)}_{t}(\hat{x}_t^i),
\end{equation}
since the state transition prior is adopted as the proposal distribution that leads to $p(x_t|y_{1,1:t-1},y_{2,1:t-1})\simeq\sum_{i=1}^N\omega_{t-1}^i\delta_{\hat{x}_t^i}$.

Since $\pi_{m,t|t}$ and $p_{m,t|t}$ have been calculated out (according to Eqns.(\ref{posterior_model_indicator}) and (\ref{eqn:posterior_m_approx}), respectively), Eqn.(\ref{eqn:bma}) can be solved by $p_{t|t}\simeq \sum_{i=1}^N\omega_t^i\delta_{\hat{x}_t^i}$, where $\omega_t^i=\sum_{m=1}^M\pi_{m,t|t}\omega_{m,t}^i, \forall i$.
Finally, A standard operation of PF, namely resampling, is performed to avoid particle degeneracy.
\section{Experiment}\label{sec:experiment}
\subsection{Experiment Setting}\label{sec:exp_setting}
We compare DMA with a single model (defined by Eqns.(1)-(3)) based benchmark PF, a static model averaging based method (SMA), and a two-stage approach (TS). The experimental data is collected from a simulated experiment for 2D target tracking (2DTT). SMA runs 2 PFs based on 2 SSMs, each corresponding to a modality, respectively. The outputs of these 2 PFs are averaged as the algorithm's output. TS follows the mainstream idea of first-detect-failure-then-do-fusion in the literature (see e.g., \cite{kumar2007method,kumar2006generalized}). It employs a single PF and a modified likelihood function $L_t(\textbf{x}_t)=(L_{1,t}(\textbf{x}_t))^{1-\alpha_{1,t}}(L_{2,t}(\textbf{x}_t))^{1-\alpha_{2,t}}$, where $\alpha_{i,t}$ denotes the estimated failure probability of the $i$th modality at time $t$, $i=1,2$.

The hidden state to be estimated is $\textbf{x}=[v_x, v_y, d_x, d_y]^T$; the 2 observation modalities considered are the angle $\theta\triangleq\arctan(d_x/d_y)$ and the range $r\triangleq\sqrt{d_x^2+d_y^2}$. Here $v$ and $d$ denote the velocity and the distance, respectively, of the target relative to the observer, the symbol $^T$ denotes transposition, and $x$ ($y$) in the subscript denotes the $X$ ($Y$) coordinate of the 2D space. In the experiment, the true model to generate normal observations in the absence of modality failures is $\textbf{x}_t\sim\mathcal{N}(\cdot|A\textbf{x}_{t-1},Q)$, $\textbf{y}_{1,t}\sim\mathcal{N}(\cdot|\arctan(d_{x,t}/d_{y,t}),\sigma_a^2)$, $\textbf{y}_{2,t}\sim\mathcal{N}(\cdot|\sqrt{d_{x,t}^2+d_{y,t}^2},\sigma_r^2)$, where
\begin{equation*}
A=
\begin{bmatrix}
1 & 0 & 0 & 0 \\
0 & 1 & 0 & 0 \\
1 & 0 & 1 & 0 \\
0 & 1 & 0 & 1 \\
\end{bmatrix},
Q=
\begin{bmatrix}
1 & 0 & 0 & 0 \\
0 & 1 & 0 & 0 \\
0 & 0 & 10 & 0 \\
0 & 0 & 0 & 10 \\
\end{bmatrix},
\end{equation*}
$\mathcal{N}(\cdot|b,B)$ denotes Gaussian distributed with mean $b$ and covariance $B$, $\sigma_a=0.1$ ($\sigma_r=1$) the standard error of the observation noise associated with the angle (range) modality, and $\sim$ means ``is drawn from".

To comprehensively evaluate the performance of the algorithm, we design 4 typical scenarios presented below. Each scenario involves a period of 300 time steps.
\begin{itemize}
\item Scenario 1. No modality failure happens at any time step.
\item Scenario 2. Modality failures happen but the 2 modalities never fail at the same time. Specifically, during the period $190\leq t\leq210$, the 1st modality fails with a probability 100\%; then for $220\leq t\leq230$, it fails with a probability 80\%. During $235\leq t\leq245$, the 2nd modality fails with a probability 100\%; then for $250\leq t\leq260$, it fails with a probability 80\%. No modality failure happens at the other time steps.
\item Scenario 3. Observations are lost at some time steps. Specifically, during $190\leq t\leq200$, observations of the 1st modality are lost; During $250\leq t\leq260$, observations of the 2nd modality are lost. No modality failure happens at the other time steps.
\item Scenario 4. During $190\leq t\leq200$ and $250\leq t\leq260$, both modalities fail with a probability 100\%. During $210\leq t\leq240$, both modalities fail with a probability 80\%. No modality failure happens at the other time steps.
\end{itemize}
Note that the above information is the ground truth that the algorithm agent is not aware of.

\begin{table*}\centering
\caption{Performance comparison in terms of averaged RMSE over 100 independent runs of each algorithm. The figure in the bracket denotes the corresponding variance.}
\begin{tabular}{c|c|c|c|c}
\hline %
       & PF               & TS              & SMA               & DMA\\\hline
Scenario 1 & 16.24 (0.002)     & 17.84 (0.011)   &18.09 (0.007)      & 16.25 (0.002)\\\hline
Scenario 2 & 136.22 (4840.381) & 19.10 (0.089)   & 51.73 (175.026)   & 22.40 (0.304)\\\hline
Scenario 3 & 16.21 (0.003)     & 17.80 (0.013)   & 24.86 (42.434)    & 16.28 (0.003)\\\hline
Scenario 4 & 149.17 (2060.894) & 412.37 (28.639) & 374.77 (5507.966) & 50.44 (5.282)\\\hline
averaged over the above scenarios & 79.46 (1725.320)  & 116.78 (7.188)  & 117.36 (1434.358) & 26.34 (1.398)\\\hline
computing time per run (in second)      & 25.71 (0.131)     & 41.68 (0.262)   & 27.49 (0.198)     & 27.71 (0.602)\\\hline
\end{tabular}
\label{Table:performance_compare}
\end{table*}
In all our experiments, the value spaces for the angle and the range modalities are set respectively as $[-\pi,\pi]$ and $[0, 1e4]$. The corresponding null likelihoods $L_1^0$ and $L_2^0$ are therefore $1/(2\pi)$ and $1/1e4$, respectively.
\subsection{Experimental Results}
Under the above experimental setting, we run the involved algorithms 100 times independently per each.
For each run, they are initialized with the same set of equally weighted particles with particle size $N=10,000$ and the same resampling operation (residual resampling is used here). Besides, all algorithms see totally the same observations.

We record the averaged root mean squared error (RMSE) over the 100 independent runs for each algorithm. The result is presented in Table I. As is shown, our DMA algorithm performs consistently well for all 4 scenarios considered, while each of its competitor methods performs unsatisfactorily in at least one scenario. Specifically, PF and SMA perform badly in Scenarios 2 and 4 and TS in Scenario 4. DMA performs especially better than all the others in Scenario 4; performs as well as PF in Scenario 1 and better than the others; slightly worse than TS in Scenario 2 and remarkably better than the others. All algorithms perform comparatively well in Scenario 3. Averaging over these 4 scenarios, we observe a starkly better performance of DMA compared with the others, as shown in the 5th line of Table I. We record the computing time of each algorithm over the Monte Carlo tests and present their mean and variance in the last line of Table I. As is shown, the computational complexity of DMA is just slightly greater than PF. This is because DMA invokes operations of generating and resampling new particles that are the same as in PF. These two operations constitute the major computing part for both algorithms based on the model considered here.

\begin{figure}[!htbp]
\centering
\includegraphics[width=0.85\linewidth]{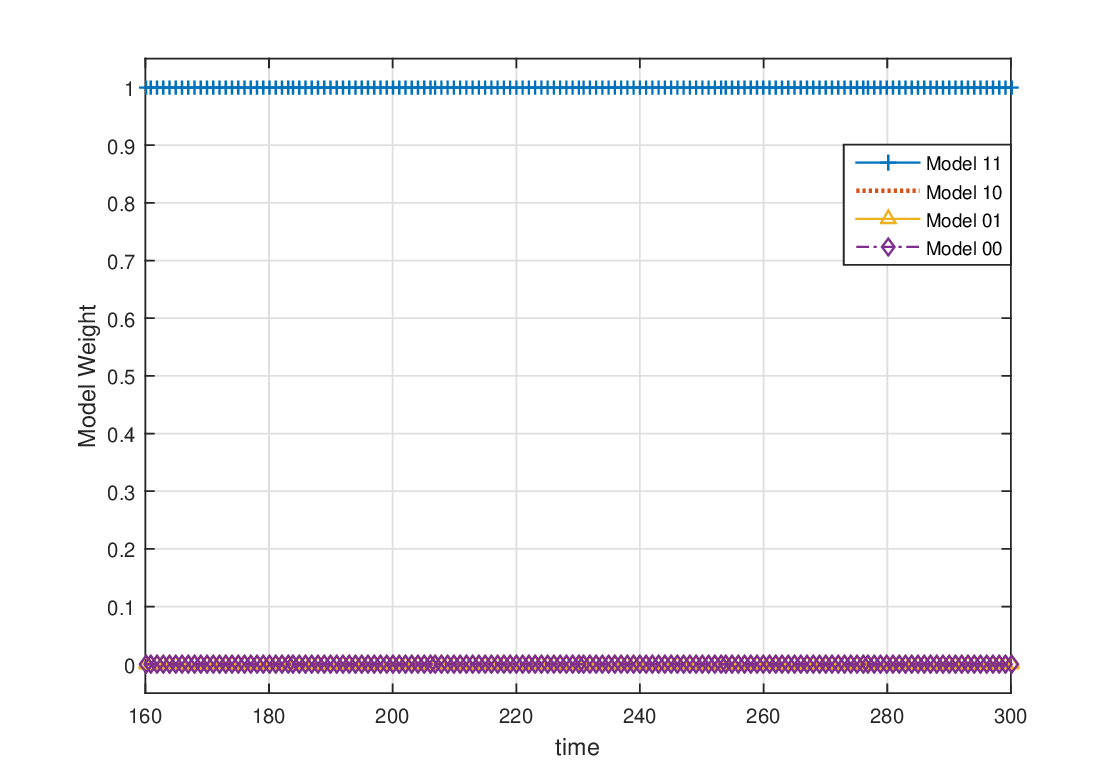}\\
\includegraphics[width=0.85\linewidth]{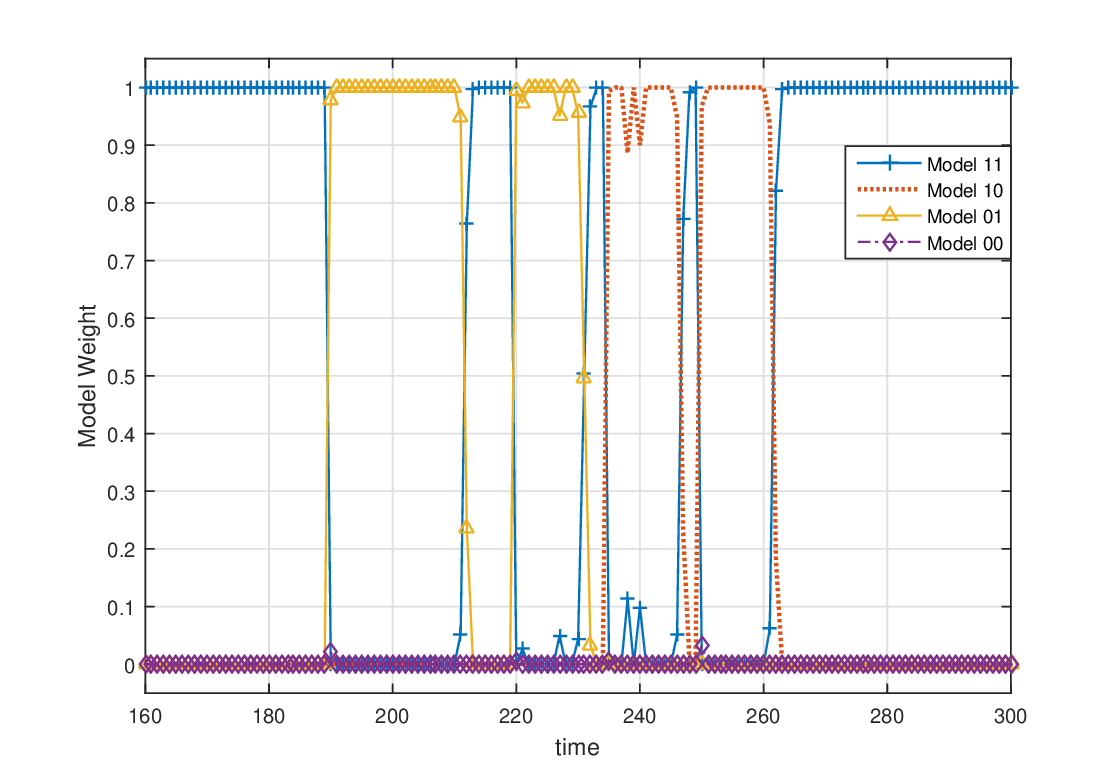}\\
\includegraphics[width=0.85\linewidth]{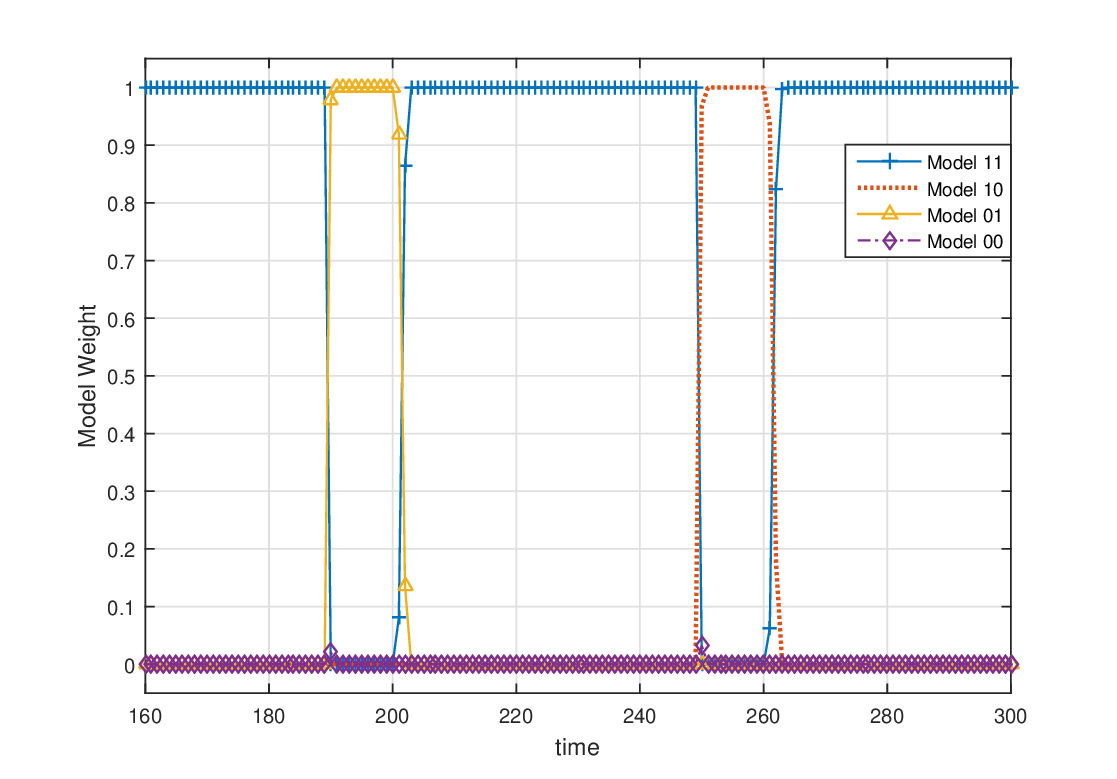}\\
\includegraphics[width=0.85\linewidth]{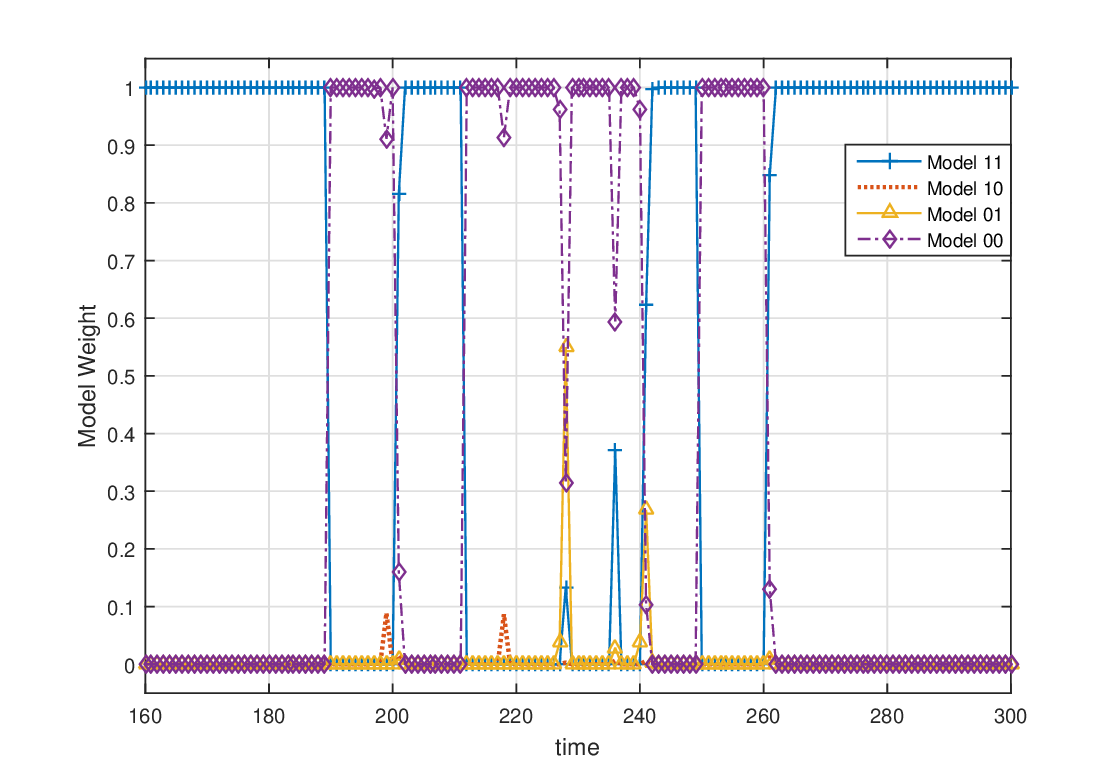}
\caption{Averaged candidate model weights given by the DMA algorithm over 100 independent runs of it. The 4 subgraphs from top to bottom correspond to Scenarios 1 to 4 respectively. ``Model 11", ``Model 10", ``Model 01" and ``Model 00" denote $\mathcal{M}_1, \mathcal{M}_2, \mathcal{M}_3$ and $\mathcal{M}_4$, respectively, defined in our model presented in subsection \ref{sec:proposed_model}. In another word, ``Model ij" represent the model in which ``\emph{usefulness}" values of the 2 modalities are $i$ and $j$, respectively.}\label{fig:model_weight}
\end{figure}

\begin{figure}[!htbp]
\centering
\includegraphics[width=0.8\linewidth]{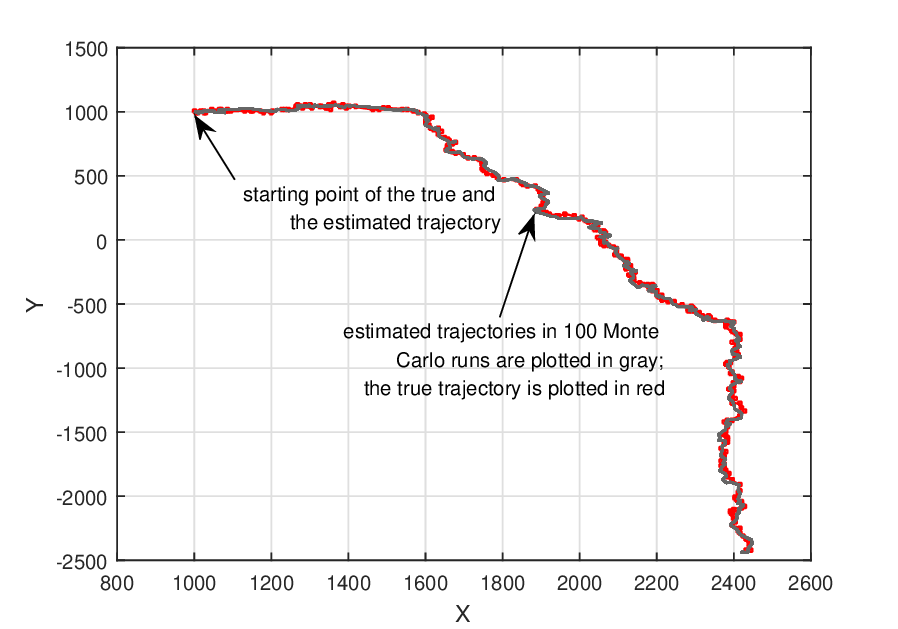}\\
\includegraphics[width=0.8\linewidth]{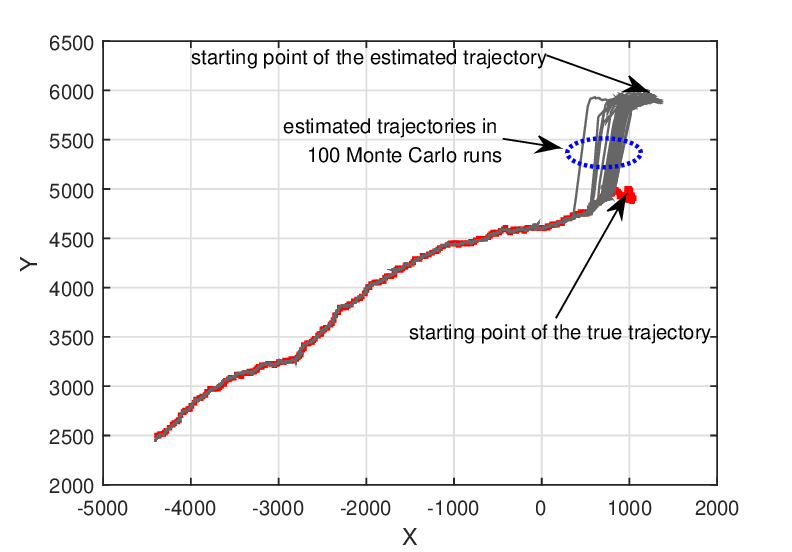}
\caption{The estimates of the target's trajectory obtained from running our DMA algorithm 100 times independently. The 2 subgraphs correspond to 2 cases with different target trajectories. For the 1st case, the algorithm is initialized accurately; for the other one, it is initialized with serious biases in the state estimation, especially in $d_y$.}\label{fig:traj}
\end{figure}

We record the candidate model weights calculated out by DMA to check if it can find the ideal way to fuse the multi-modal observations. Comparing the ground truth of the modality failures, which is presented in subsection \ref{sec:exp_setting}, with the result shown in Fig.\ref{fig:model_weight}, we see that DMA can always infer out accurately which modality fails at when for all scenarios considered. In another word, DMA can always adapt to modality failures and find the best way to fuse the observations.
Finally, we check the influence of the target trajectory and the initialization bias on its performance in Fig.\ref{fig:traj}. It is shown that DMA performs consistently well for different trajectories and converges fast in case of initialization bias.
\section{Conclusions}
The issue of DF has been investigated a lot in the literature, while robust real-time cross-modal DF in nonlinear non-Gaussian dynamic processes remains a challenge. To address it, this letter presented a novel model uncertainty perspective and an efficient algorithm derived based on this perspective. Experimental results showed that the proposed approach performed remarkably better than state-of-the-art methods, while its complexity was comparable to a conventional PF algorithm. Here the system dynamics is used as the proposal for generating new particles. It is potential to consider alternative proposals, such as those discussed in \cite{li2017particle,herbst2019tempered,liu2008particle,gustafsson2013marginal,van2000unscented,del2006sequential,devlin2021no}, for improving our approach. Although the advantage of using those proposals has been demonstrated in single-model-based contexts, it is interesting to explore their potential benefits in the multi-model setting presented here.

\bibliographystyle{IEEEbib}
\bibliography{mybibfile}
\end{document}